\documentclass[nofootinbib,superscriptaddress,twocolumn]{revtex4-1}

\usepackage{amsmath,amsfonts,amssymb}
\usepackage{wrapfig}
\usepackage{graphicx}
\usepackage{bbm}
\usepackage{color}
\usepackage{float}
\usepackage{subfigure}

\usepackage{textcomp}
\usepackage[british]{babel}
\usepackage{pifont}

\usepackage[dvipsnames]{xcolor}
\usepackage[colorlinks=true,linkcolor=blue,citecolor=blue]{hyperref}

\makeatother

\begin{document}

\title{Projective simulation with generalization}

\author{Alexey A. Melnikov}
\email[Correspondence to: ]{alexey.melnikov@uibk.ac.at}
\affiliation{Institute for Theoretical Physics, University of Innsbruck, Technikerstra{\ss}e 21a, 6020 Innsbruck, Austria}
\affiliation{Institute for Quantum Optics and Quantum Information, Austrian Academy of Sciences, Technikerstra{\ss }e 21a, 6020 Innsbruck, Austria}

\author{Adi Makmal}
\affiliation{Institute for Theoretical Physics, University of Innsbruck, Technikerstra{\ss}e 21a, 6020 Innsbruck, Austria}

\author{Vedran Dunjko}
\affiliation{Institute for Theoretical Physics, University of Innsbruck, Technikerstra{\ss}e 21a, 6020 Innsbruck, Austria}

\author{Hans J. Briegel}
\affiliation{Institute for Theoretical Physics, University of Innsbruck, Technikerstra{\ss}e 21a, 6020 Innsbruck, Austria}
\affiliation{Department of Philosophy, University of Konstanz, Fach 17, 78457 Konstanz, Germany}

\begin{abstract}
The ability to generalize is an important feature of any intelligent agent. Not only because it may allow the agent to cope with large amounts of data, but also because in some environments, an agent with no generalization capabilities cannot learn. In this work we outline several criteria for generalization, and present a dynamic and autonomous machinery that enables projective simulation agents to meaningfully generalize. Projective simulation, a novel, physical approach to artificial intelligence, was recently shown to perform well in standard reinforcement learning problems, with applications in advanced robotics as well as quantum experiments. Both the basic projective simulation model and the presented generalization machinery are based on very simple principles. This allows us to provide a full analytical analysis of the agent's performance and to illustrate the benefit the agent gains by generalizing. Specifically, we show that already in basic (but extreme) environments, learning without generalization may be impossible, and demonstrate how the presented generalization machinery enables the projective simulation agent to learn.
\end{abstract}

\flushbottom
\maketitle

\thispagestyle{empty}

\section*{Introduction}

The ability to act upon a new stimulus, based on previous experience with similar, but distinct, stimuli, sometimes denoted as \emph{generalization}, is used extensively in our daily life. As a simple example, consider a driver's response to traffic lights: The driver need not recognize the details of a particular traffic light in order to respond to it correctly, even though traffic lights may appear different from one another. The only property that matters is the color, whereas neither shape nor size should play any role in the driver's reaction. Learning how to react to traffic lights thus involves an aspect of generalization. 

A learning agent, capable of a meaningful and useful generalization is expected to have the following characteristics: 
(a) an ability for \emph{categorization} (recognizing that all red signals have a common property, which we can refer to as \emph{redness}); 
(b) an ability to \emph{classify} (a new red object is to be related to the group of objects with the redness property);
(c) ideally, only generalizations that are \emph{relevant} for the success of the agent should be learned (red signals should be treated the same, whereas {square-shaped signals} should not, as they share no property that is of relevance in this context); (d) correct actions should be \emph{associated} with relevant generalized properties (the driver should stop whenever a red signal is shown);  and (e) the generalization mechanism should be \emph{flexible}.

To illustrate what we mean by ``flexible generalization", let us go back to our driver. After learning how to handle traffic lights correctly, the driver tries to follow arrow signs to, say, a nearby airport. Clearly, it is now the shape category of the signal that should guide the driver, rather than the color category. The situation would be even more confusing, if the traffic signalization would suddenly be based on the shape category alone: square lights mean ``stop" whereas circle lights mean ``drive". To adapt to such environmental changes the driver has to give up the old color-based generalization and build up a new, shape-based, generalization.  Generalizations must therefore be flexible. 

Refs.~\cite{Holland_etal_Induction86,2013_Saitta_book} provide a broad account of generalization in artificial intelligence. In reinforcement learning (RL), where an agent learns via interaction with a rewarding environment~\cite{barto1998reinforcement,russell2010artificial,RL_2012_book}, generalization is often used as a technique to reduce the size of the percept space, which is potentially very large. Two useful recent summaries can be found in Refs.~\cite{2008_van_Otterlo, 2010_Ponsen}. For example, in the Q-learning~\cite{Watkins_1989} and SARSA~\cite{rummery1994line} algorithms, it is common to use function approximation methods~\cite{barto1998reinforcement,melo2008analysis,russell2010artificial,RL_2012_book}, realized by e.g.\ tile coding (CMAC)~\cite{1975_Albus_CMAC,sutton1996generalization}, neural networks~\cite{boyan1995generalization, whiteson2006evolutionary, 2015_QL_Nature}, decision trees~\cite{Pyeatt98decisiontree, 2005_Q_iteration}, constructive function approximation~\cite{Utgoff98constructive}, or support vector machines~\cite{1995_Vapnik,2007_Laumonier,russell2010artificial}, to implement a generalization mechanism. 
Alternatively, in learning classifier systems (LCS), generalization is facilitated by using the wildcard \# character, which, roughly speaking, means that a particular category is irrelevant for the present task environment~\cite{Holland76,Holland86,2009_LCS_Review}. This relates to recent works on state abstraction, which maps different states to which the agent should react similarly, to a single, abstract one~\cite{2005_State_abstraction,2006_State_abstraction,2011_State_abstraction}. Temporal abstraction or grouping of actions is addressed in what is known as hierarchical-RL \cite{Sutton1999181, 2012_HRL}. These methods are usually applied to problems where the states and/or actions are factored into categories (such as, e.g., factored Markov decision processes~\cite{2008_van_Otterlo}). More modern approaches to generalization include relational-RL \cite{2004_Tadepalli_relationalRL,2008_van_Otterlo} and feature-RL~\cite{hutter2009feature, 2012_Nguyen, daswani2014feature}.

In this work we contribute to the development of the recently introduced model of projective simulation (PS)~\cite{briegel2012projective}, by exhibiting an approach for generalization within the PS framework. PS is a physical approach to artificial intelligence which is based on stochastic processing of experience. It uses a particular type of memory, denoted as \emph{episodic \& compositional memory} (ECM), which is structured as a directed, weighted network of \emph{clips}, where each clip represents a remembered percept, action, or sequences thereof. Once a percept is observed, the network is activated, invoking a random walk between the clips, until an action clip is hit and couples out as a real action of the agent.  

The generalization process within PS is achieved in this work by, roughly speaking, the generation of abstracted clips, which represent commonalities between percepts, or more precisely, subsets of the percept space. These, in turn, influence behavior in new situations based on similar previous experiences. The method we introduce is a step toward more advanced approaches to generalization within PS and is suitable for medium-scale task environments. In more complicated environments, the large number of possible abstractions may harm the performance of the agent. We address the question how this could be combated in the discussion section.

The PS approach to artificial intelligence, arguably, stands out as promising from different perspectives: First, random walks, which constitute the basic internal dynamics of the model, have been well-studied in the context of randomized algorithm theory~\cite{Randomized_Algorithms_1995} and probability theory, thus providing an extensive theoretical toolbox for analyzing related models;  second, the PS model, by design, represents a stochastic physical system which points to possible physical (rather than computational) realizations. This relates PS to the framework of embodied artificial agents~\cite{Understanding_Intelligence_1999}; last, the physics aspects of the model offer a route toward the research into quantum-enhanced variants of PS: the underlying random walk was already shown to naturally extend to a quantum many-body master equation~\cite{briegel2012projective}. Related to this, the fact that the deliberation of the agent centers around a random walk process, opens up a route for advances by using quantum random walks instead. 
In quantum random walks, roughly speaking, the probabilistic re-positioning of the walker is replaced by a \emph{quantum superposition} of moves, by switching from a stochastic to a coherent quantum dynamics. This allows one to exploit quintessential quantum phenomena, including quantum interference and quantum parallelism. Quantum walks have been increasingly more employed in recent times as a new framework for the development of new quantum algorithms. Over the course of the last decade, polynomial and exponential improvements in computational complexity have been reported, over classical counterparts~\cite{Childs:2003,Kempe,KMOR}.
Utilizing this methodology in the context of reinforcement learning, it was recently shown, by some of the authors and collaborators, that a quantum variant of the PS agent exhibits a quadratic speed-up in deliberation time over its classical analogue, which leads to a similar speed-up of learning time in active learning scenarios~\cite{paparo2014quantum,dunjko2015quantum,friis2015coherent,PhysRevLett.QML}. This quantum advantage in the decision-making process of the quantum PS agent was recently experimentally demonstrated using a small-scale quantum information processor based on trapped ions~\cite{sriarunothai2017speedingup}.

In the PS model, learning is realized by internal modification of the clip network, both in terms of its structure and the weights of its edges. Through interactions with a rewarding environment, the clip network adjusts itself dynamically, so as to increase the probability of performing better in subsequent time steps (see below, for a more detailed description of the model). Learning is thus based on a ``trial and error" approach, making the PS model especially suitable for solving RL tasks. Indeed, recent studies showed that the PS agent can perform very well in comparison to standard models, in both basic RL problems~\cite{mautner2013projective} and in standard benchmark tasks, such as the ``grid world" and the continuous-domain ``mountain car problem"~\cite{melnikov2014projective}. Due to the flexibility of the PS framework it can also be used in contexts beyond textbook RL. Recent applications are, for instance, in the problem of learning complex haptic manipulation skills~\cite{simon2016} and in the problem of learning to design complex quantum experiments~\cite{melnikov2017}.

Here we present a simple dynamical mechanism which allows the PS network to evolve, through experience, to a network that represents and exploits similarities in the perceived percepts, i.e.\ to a network that can generalize. Using such a network we address the problem of RL in task environments, which require some aspects of generalization (function approximation) to be solved. Standard approaches to such task environments rely on external machinery, a function approximator, which then has to be combined to otherwise ``raw", tabular, RL machinery, e.g. temporal difference learning model-free methods such as Q-learning and SARSA. In contrast, to achieve the same goal, here we use the more elaborate structure of the PS model, which is not represented by a table, but by a directed graph, rather than external machinery, e.g., such as in the case of Q-learning~\cite{melo2008analysis,2015_QL_Nature}. Naturally, our proposed machinery also internally realizes a function approximator, but its structure and updates arise from the very basic learning rules of the PS model. The generalization mechanism, which is inspired by the wildcard notion of LCS~\cite{Holland76,Holland86,2009_LCS_Review}, is based on a process of abstraction which is systematic, autonomous, and, most importantly, requires no explicit prior knowledge of the agent. This is in contrast with common RL models with a function approximator, which often require additional prior knowledge in terms of an additional input~\cite{2010_Ponsen}.
Moreover, we show that once the PS agent is provided with this machinery which allows it to both \emph{categorize} and \emph{classify}, the rest of the expected characteristics we listed above follow directly from the basic learning rules of the PS agent. In particular, we show that \emph{relevant} generalizations are learned, that the agent \emph{associates} correct actions to generalized properties, and that the entire generalization scheme is \emph{flexible}, as required.  

PS with generalization, in comparison with other solutions, does not rely on external machinery. Instead, it is a homogeneous approach where the generalization mechanism is based on the basic PS principles, in this case specifically the dynamic growth of the clip network. One can see several advantages to our approach: First, it allows for a relatively straightforward theoretical treatment of the performance, including analytic formulas characterizing the performance of generalization and learning. Second, we do not rely on powerful classifying machinery which can significantly increase the model complexity of the agent (in particular if neural networks are utilized), which may be undesirable. 
Finally, and for our agenda very relevant, sticking to just the basic random walk mechanism offers a natural and systematic route to quantization of the overall dynamics. As we have mentioned earlier, this can lead to improvements in computational complexity and, in principle, also in space complexity of the model. In contrast, for heterogeneous approaches, e.g. Q-learning combined with a neural network, there exist no clear routes for useful quantization, and no firm results proving improvements have been established for the quantization of either, let alone for a combination.

While in most RL literature elements of generalization are considered as means of tackling the ``curse of dimensionality"~\cite{RL_2012_book}, as coined by Bellman~\cite{bellman1957dynamic} and discussed above, they are also strictly necessary for an agent to learn in certain environments~\cite{barto1998reinforcement}. Here we consider a type of environment where, irrespective of its available resources, an agent with no generalization ability cannot learn, i.e. it performs no better than a fully random agent.

Following this, we show that the PS model, when enhanced with the generalization mechanism, is capable of learning in such an environment. Along numerical illustrations we provide a detailed analytical description of the agent's performance, with respect to its success- and learning-rates (defined below). Such an analysis is feasible due to the simplicity of the PS model, both in terms of the number of its free parameters, and its underlying equations (see also Ref.~\cite{mautner2013projective}), a property we extensively exploit. The main contribution of this paper is thus to demonstrate how the inherent features of the PS model can be used to solve RL problems that require nontrivial notions of generalization, importantly without relying on external classifier machinery. While it is also possible to sacrifice homogeneity, and combine PS with external machinery (in which case a direct comparison between the PS and other RL models both enhanced by external machinery would be warranted), in this work we strive to develop the theory of the PS model on its own.

\section*{Results}

The remainder of this paper is structured as follows. We first begin, for completeness, with a description of the PS model and a formal comparison of the PS model to standard RL techniques. We then present the proposed generalization mechanism, examine its performance in a simple case and illustrate how it gives rise to a meaningful generalization, as defined above. 
Next, we study the central scenario of this paper, in which generalization is an absolute condition for learning. After describing the scenario and showing that the PS agent can cope with it, we analyze its performance analytically. 
Finally, we study this scenario for an arbitrary number of categories, and observe that the more there is to categorize the more beneficial is the proposed mechanism.

\subsection*{The PS model}

In what follows we shortly summarize the basic principles of the PS model, for  more detailed descriptions we refer the reader to Refs.~\cite{briegel2012projective,mautner2013projective,melnikov2014projective,makmal2016meta}. 

\begin{figure}[h]
	\centering
	\includegraphics[width=8cm]{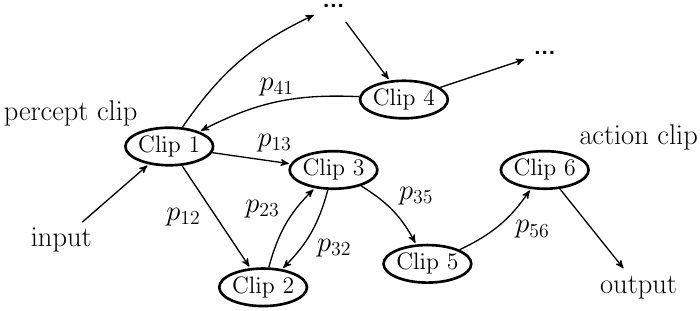}
	\caption{The PS clip network.}
	\label{fig:Fig1}
\end{figure}

The central component of the PS agent is the so-called clip network, which can, abstractly, be represented as a directed graph, where each node is a clip,  and directed edges represent allowed transitions, as depicted in Fig.~\ref{fig:Fig1}. 
Whenever the PS agent perceives an input, the corresponding percept clip is excited (e.g.\ Clip~1 in Fig.~\ref{fig:Fig1}). This excitation marks the beginning of a random walk between the clips until an action clip is hit (e.g.\ Clip~6 in Fig.~\ref{fig:Fig1}), and the corresponding action is performed. The random walk is carried out according to time-dependent probabilities $p_{ij}$ to hop from one node to another.

Formally, \emph{percept} clips are defined as $K$-tuples $s=(s_1,s_2,...,s_K) \in {\cal S} \equiv {\cal S}_1 \times {\cal S}_2 \times ... \times {\cal S}_K,s_i \in \{1,...,|{\cal S}_i|\}$, where $|{\cal S}| = |{\cal S}_1| \cdots |{\cal S}_K|$ is the number of possible percepts. Each dimension may account for a different type of perceptual input such as audio, visual, or sensational, where the exact specification (number of dimensions $K$ and the perceptual type of each dimension) and resolution (the size $|{\cal S}_i|$ of each dimension) depend on the physical realization of the agent. In what follows, we regard each of the $K$ dimensions as a different category. \emph{Action} clips are similarly given as $M$-tuples: $a=(a_1,a_2,...,a_M) \in {\cal A} \equiv {\cal A}_1 \times {\cal A}_2 \times ... \times {\cal A}_M,a_i \in \{1,...,|{\cal A}_i|\}$, where $|{\cal A}| = |{\cal A}_1| \cdots |{\cal A}_M|$ is the number of possible actions. Once again, each of the $M$ dimensions provides a different aspect of an action, e.g.\ walking, jumping, picking-up, etc. Here, however, we restrict our analysis to the case of $M=1$ and varying $|{\cal A}_1|$.

Each directed edge from clip $c_i$ to clip $c_j$ has a time dependent weight $h^{(t)}(c_i,c_j)$, which we call the $h$-value. The $h$-values define the conditional probabilities of hopping from clip $c_i$ to clip $c_j$ according to
\begin{equation}
  p^{(t)}(c_j|c_i) = \dfrac{h^{(t)}(c_i,c_j)}{\sum_{k} h^{(t)}(c_i,c_k)}.
\label{eq:probablititesBasic}
\end{equation}
At the beginning, all $h$-values are initialized to the same fixed value $h_0=1$. This ensures that, initially, the probability to hop from any clip to any of its neighbors is completely uniform.
The conditional probabilities defined by Eq.~(\ref{eq:probablititesBasic}) will be used throughout the paper unless stated otherwise. One can also define a different function for conditional probabilities known as the softmax function
\begin{equation}
  p^{(t)}(c_j|c_i) = \frac{\mathrm{e}^{\beta h^{(t)}(c_i,c_j)}}{\sum_{k} \mathrm{e}^{\beta h^{(t)}(c_i,c_k)}},
\label{eq:probablititesSoftmax}
\end{equation}
where $\beta$ is an inverse temperature parameter, the lower the temperature -- the higher the chance to traverse an edge with the largest $h$-value.

Learning takes place by the dynamical strengthening and weakening of the internal $h$-values, in correspondence to an external feedback, i.e.\ a reward $\lambda$, coming from the environment. 
Specifically, the update of the $h$-values is done according to the following update rule: 
\begin{eqnarray}
h^{(t+1)}(c_i,c_j)&=&h^{(t)}(c_i,c_j) - \gamma (h^{(t)}(c_i,c_j) - 1)\nonumber\\
&+&\sum_l\delta(c_i,c_{k_l})\delta(c_j,c_{m_l})\lambda^{(t+1)},
\label{eq:hupdate}
\end{eqnarray}
where the reward $\lambda$ is non-negative ($\lambda=0$ implies no reward), and is added only to the $h$-values of the edges $(c_{k_l},c_{m_l})$ that were traversed in the last random walk. The update rule can also handle negative rewards given that the probability function is defined by Eq.~(\ref{eq:probablititesSoftmax}) so that the transition probabilities $p(c_j|c_i)$ are guaranteed to remain non-negative. The damping parameter $0 \leq \gamma \leq 1$ weakens the $h$-values of all edges and allows the agent to forget its previous experience, an important feature in changing environments~\cite{briegel2012projective,mautner2013projective,makmal2016meta}. The damping term in Eq.~(\ref{eq:hupdate}) with a nonzero $\gamma$ is however not needed in stationary environments, such as the contextual bandit task~\cite{Bandits05}. In the contextual bandit task the PS agent with the update rule in Eq.~(\ref{eq:hupdate}) with $\gamma=0$ realizes the optimal policy in the limit. Note that although the update rule in Eq.~(\ref{eq:hupdate}) does not have an explicit tunable learning rate, the effective separation between different $h$-values are tunable by changing the inverse temperature parameter $\beta$ in the probability function.

In order to cope with more general environments, which may include temporal correlations, for instance through delayed rewards, the PS model utilizes the so-called glow mechanism~\cite{mautner2013projective}. In this mechanism, to each edge an additional variable $g(c_i,c_j)\in [0,1 ]$ is assigned. The edge $g$-value is set to 1 whenever an edge is traversed, and for each time-step that the edge was not used, it dissipates according to the rule
\begin{equation}
g^{(t+1)}(c_i,c_j) = (1-\eta)~g^{(t)}(c_i,c_j),
\label{eq:gupdate}
\end{equation}
where the rate $\eta \in [0,1 ]$ is a parameter of the model. The $h-$value update rule for the non-traversed edges, with glow, assumes the form
\begin{eqnarray}
h^{(t+1)}(c_i,c_j)&=&h^{(t)}(c_i,c_j) - \gamma (h^{(t)}(c_i,c_j) - 1)\nonumber\\
&+&g^{(t+1)}(c_i,c_j)\lambda^{(t+1)}.
\label{eq:hupdateglow}
\end{eqnarray}
To clarify, with glow, the edges whose transition did not result in obtaining a reward still obtain a fraction of a later issued reward proportional to the current glow value. The latter corresponds to how far in the past, relative to the later rewarded time-step, the particular edge was used. The glow mechanism thus allows for a future reward to propagate back to previously used edges, and enables the agent to perform well also in settings where the reward is delayed (e.g. in the grid world and the mountain car tasks~\cite{melnikov2014projective}) and/or is contingent on more than just the immediate history of agent-environment interaction, such as in the n-ship game, as presented in Ref.~\cite{mautner2013projective}.
The described learning mechanisms specified by Eq.~(\ref{eq:gupdate})-(\ref{eq:hupdateglow}) fully define the basic PS agent with fixed learning parameters $\gamma$ and $\eta$. The values of these learning parameters, or meta-parameters, have to be set properly such that the agent performs optimally in a certain task. However, as it was shown in Ref.~\cite{makmal2016meta}, the PS model can naturally be extended to account for self-monitoring of its own meta-parameters and one and the same agent can reach near-optimal to optimal success rates in different kinds of reinforcement learning problems.

In this work, we present extensions which make use of the capacity of the PS model to generate new clips dynamically, for the purpose of generalization. To focus our study of  the performance of such a PS agent, here we introduce simplest environmental scenarios which highlight the critical aspects of generalization, and analyze how the PS with generalization performs. While these simple settings all fit into the contextual bandit framework \cite{Bandits05} and hence do not require glow, the same mechanism is of course readily applied to more complex task environments (e.g. with temporal correlations) and analogous generalization results will hold. In the next section we will briefly reflect on the relation of the basic PS model (without generalization) with the more standard reinforcement learning machinery, which will further put this work into context.

\subsubsection*{Projective simulation and reinforcement learning}

The basic PS model can be viewed as an explicitly reward-driven RL model. Unlike most standard RL models, it does not include an explicit approximation of the state-value or the action-value function. As a consequence of this, PS is also simpler in the sense that the value function and the policy are not optimized over separately, but rather the optimization occurs concurrently. Despite these structural differences, the PS model can be related to other standard RL approaches. Quantitively, the PS model was shown to perform similar in comparison to the standard RL models in benchmark tasks~\cite{melnikov2014projective,mautner2013projective,bjerland2015projective}. In addition to this quantitative relationship, a formal relationship between the $h$-value matrix and the action-value Q-matrix can be derived on a formal level, from which many fundamental properties of RL algorithms are qualitatively recovered in PS as well. For instance, in the setting of stationary environments with immediate rewards, the update rule of the basic two-layered PS model, given percept $s$ and action $a$ reads as
\begin{equation}
h^{(t+1)}(s,a) = h^{(t)}(s,a) + \lambda^{(t+1)}.
\end{equation}
It is clear that the $h-$values, when normalized by the number of realized transitions over the history, converge to the (immediate) value of this transition. Moreover, given the policy-generating rule given in Eq.~(\ref{eq:probablititesBasic}), it holds that the probability of outputting the action of the most rewarded transition converges to unity. This is equivalent to employing a greedy policy over a converged action-value function in standard RL, hence also implies that this basic PS handles stationary contextual bandit problems. Nonetheless, the $h-$values in PS are not meant to represent action values, but actually stem from descriptions of physical dynamics (as coupling coefficients). Going a few steps further, in the setting of more general MDP environments (with delayed rewards) we can contrast Eq.~(\ref{eq:hupdateglow}) of the basic PS to the standard SARSA Q-matrix update~\cite{rummery1994line,barto1998reinforcement}. For the latter we have the expression
\begin{eqnarray}
Q^{(t+1)}(s,a)&=&Q^{(t)}(s,a) + \alpha \big[\lambda^{(t+1)} + \gamma_{RL} Q^{(t)}(s', a')\nonumber\\
&-&Q^{(t)}(s,a)\big],
\end{eqnarray}
where the discounted value of the next state $s'$ and action $a'$, $\gamma_{RL} Q(s', a')$ (note, this $\gamma_{RL}$ does not correspond to the $\gamma$ term in the basic PS, but constitutes the discount factor), ensures the current action value obtains a fraction of the value of the subsequently realized step.

In PS, the update rule is represented in Eq.~(\ref{eq:gupdate}) and (\ref{eq:hupdateglow}). Note that this edge $h$-value is updated regardless whether this edge was traversed, that is, in each time-step. On the other hand, this $h-$value plays an active role in the outputs of the PS agent only in some subsequent time-step $t''$ when a transition from clip $c_i$ is required. In this interval, from $t'$ when the edge $(c_i,c_j)$ was traversed last, until $t''$ when the clip $c_i$ is encountered again, this edge accumulates discounted rewards. At time step $t''$ when this particular edge may be used again, the relevant $h$-value reads as
\begin{equation}
  h^{(t'')}(c_i,c_j) = h^{(t')}(c_i,c_j) + \lambda^{(t'+1)} + \sum_{k=t'+2}^{t''} (1-\eta)^{t''-k} \lambda^{(k)},
\end{equation}
where we have set the damping $\gamma$ term to zero to simplify the expression. The term $\sum_{k=t'+2}^{t''} (1-\eta)^{t''-k} \lambda^{(k)}$ accounts for all the future rewards which followed the $(c_i,c_j)$ transition, which occurred at time-step $t'$. The term $\sum_{k=t'+2}^{t''} (1-\eta)^{t''-k} \lambda^{(k)}$ is closely related to $\gamma_{RL} Q(s', a')$ -- the first term captures the discounted future rewards experienced by \emph{this particular agent} in its future realized steps, starting from the next step. The term $\gamma_{RL}Q(s', a'),$ captures the current approximation of what the future rewards will be (under the current policy), also starting from the next step, discounted by $\gamma_{RL}$. In other words, $\gamma_{RL}$ in SARSA plays the same functional role as the $g-$value decay rate $(1-\eta)$ in PS. To further clarify, note that $\gamma_{RL} Q(s', a')$ corresponds to the $\sum_{k=t'+2}^{t''} (1-\eta)^{t''-k} \lambda^{(k)}$ term, but computed for an averaged agent, averaged over the sequences of subsequent moves, given the agent's policy. Note, however, that in the case of PS, the edge $(c_i,c_j)$ will be traversed many times over the course of learning, leading to an effective averaging of the future-reward term $\sum_{k=t'+2}^{t''} (1-\eta)^{t''-k} \lambda^{(k)}$. In the case of non-zero damping, all the rewards in the sum would also undergo proportional damping, but this yields a complicated expression which obfuscates the general trends in behavior.

We note that  this heuristic analysis certainly does not constitute any formal statement about the relationship of the basic PS with other reinforcement learning models. While a full analysis goes beyond the scope of this paper, already this heuristic suggests that the expected performance of it should not differ, qualitatively, from other models. This has been confirmed empirically through benchmarking against other models \cite{melnikov2014projective,mautner2013projective,bjerland2015projective}, and while in some occasions PS outperformed Q-Learning or Dyna-planning, and in some it underperformed, the global trends were comparable.

\begin{figure*}[t!]
	\centering
	\includegraphics[width=16cm]{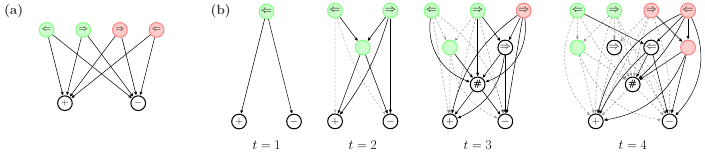}
	\caption{(a) The basic PS network as it is built up for the driver scenario. Four percept clips (arrow, color) in the first row are connected to two action clips ($+\backslash-$) in the second row. Each percept-action connection is learned independently. (b) The enhanced PS network as it is built up for the driver scenario, during the first four time steps. The following sequence of signals is shown: left-green ($t=1$), right-green ($t=2$), right-red ($t=3$), and left-red ($t=4$). Four percept clips (arrow, color) in the first row are connected to two layers of wildcard clips (first layer with a single wildcard and second layer with two) and to two action clips ($+\backslash-$) in the fourth row. Newly created edges are solid, whereas existing edges are dashed (relative weights of the $h$-values are not represented).}
	\label{fig:Fig2}
\end{figure*}

Having discussed the similarities between the basic PS model and other standard RL algorithms, now we turn our attention to the differences, which is the topic of this paper. Unlike in the mentioned RL models, where learning revolves around the estimation of value functions, in PS learning is embodied in the re-configuration of the clip network. This includes the update of transition probabilities but also the dynamical network restructuring (via e.g. clip creation~\cite{briegel2012projective} applied here for generalization). The latter has no analog in standard RL approaches we discussed previously, and only makes sense since the clip network is manifestly not a representation of value functions, but conceptually a different object. In this work we further explore this capacity of the PS model, by showing how it can be utilized to handle problems of generalization. Other possibilities, related to action fine-graining, were previously studied in Refs.~\cite{mautner2013projective, tiersch2014adaptive}. Here, the PS approach diverges from standard methodology, which, to our knowledge without exception, tackles this problem by using external machinery like classifiers (developed in the context of e.g. supervised learning), or, more generally, function approximators. We reiterate that such additional machinery could also be used with PS, but this comes at a cost which we elaborated on previously. In the next section we focus on how simple tasks which require generalization can be resolved using PS with dynamic clip generation.

\subsection*{Generalization within PS}

Generalization is usually applicable when the perceptual input is composed of more than a single category. In the framework of the PS model, this translates to the case of $K > 1$ in percept space. In particular, when two (or more) stimuli are similar, i.e.\ share a set of common features, or, more precisely, have the same values for some of the categories, it may be useful to process them in a similar way. Here we enhance the PS model with a simple but effective generalization mechanism based on this idea.

The key feature of this mechanism is the dynamical creation of a class of \emph{abstracted} clips that we call \emph{wildcard} clips. Whenever the agent encounters a new stimulus, the corresponding new percept clip is created and compared pairwise to all existing clips. For each pair of clips whose $1 \leq l \leq K$ categories carry different values, a new wildcard clip is created (if it does not already exist) with all the different $l$ values replaced with the wildcard symbol $\#$. Such a wildcard clip then represents a categorization based on the remaining $K-l$ common categories.   

A wildcard clip with $l$ wildcard symbols is placed in the $l$th layer of the clip network (we consider the percept clip layer as the zeroth layer). In general, there can be up to $K$ layers between the layer of percept clips and the layer of action clips, with $\binom{K}{l}$ wildcard clips in layer $l$ for a particular percept. From each percept- and wildcard-clip there are direct edges to all action clips and to all matching higher-level wildcard clips. By matching higher-level wildcard clips, we mean wildcard clips with more wildcard symbols, whose explicit category values match with those of the lower-lever wildcard clip. In essence, a matching higher-level wildcard clip (e.g.\ the clip $(s_1,s_2,\#,\#)$) generalizes further a lower-level wildcard clip (e.g.\ $(s_1,s_2,s_3,\#)$).

\begin{figure*}[t!]
	\centering
	\includegraphics[width=14.9cm]{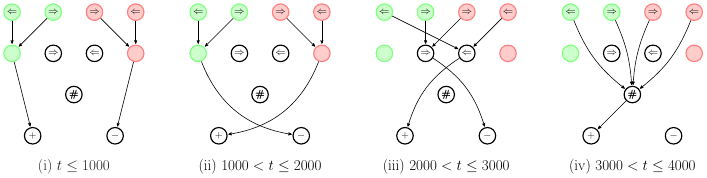}
	\caption{The enhanced PS network configurations (idealized) as built up for each of the four phases of the driver scenario (see text). Only indirect strong edges are shown. Different wildcard clips allow the network to realize different generalizations. Categorization and classification are realized by the structure of the network, whereas relevance, correctness and flexibility come about through the update rule of Eq.~(\ref{eq:hupdate}).}
	\label{fig:Fig3}
\end{figure*}

To demonstrate how this mechanism operates we consider the example from the introduction. An agent acts as a driver who should learn how to deal with traffic lights and arrow signs. While driving, the agent sees a traffic light with an arrow sign and should choose among two actions ($|{\cal A}_1|=2$): continue driving ($+$) or stop the car ($-$). The percepts that the agent perceives are composed of two categories ($K=2$): 
color and direction. Each category has two possible values ($|{\cal S}_1|=|{\cal S}_2|=2$): red and green for the color, and left and right for the direction. 
At each time step $t$ the agent thus perceives one of four possible combinations of colors and arrows, randomly chosen by an environment, and chooses one of the two possible actions. In such a setup, the basic PS agent, described in the previous section, would have a two-layered network of clips, composed of four percept clips and two action clips, as shown in Fig.~\ref{fig:Fig2}(a). It would then try to associate the correct action for each of the four percepts separately. The PS with generalization, on the other hand, has a much richer playground: it can, in addition, connect percept clips to intermediate wildcard clips, and associate wildcard clips with action clips, as we elaborate below. 

The development of the enhanced PS network is shown step by step in Fig.~\ref{fig:Fig2}(b) for the first four time steps of the driver scenario (a hypothetical order of percepts is considered for illustration). When a left-green signal is perceived at time $t=1$, the corresponding percept clip is created and connected to the two possible actions ($+\backslash-$) with an initial weight $h_0$. In the second time step $t=2$, a right-green signal is shown. This time, in addition to the creation of the corresponding percept clip, the wildcard clip (\#, green) is also created (since both of the encountered percepts are green) and placed in the first layer of the network. To simplify the visualization we draw the wildcard clip (\#, green) as a green circle with no direction (and without the $\#$ symbol). In general, we omit one $\#$ symbol in all figures. Newly created edges are shown in Fig.~\ref{fig:Fig2}(b) as solid lines, whereas all previously created edges are shown as dashed lines. Next, at time $t=3$, a right-red signal is presented. This leads to the creation of the ($\Rightarrow$, $\#$)-clip, because both the second and the third percepts have a right arrow. Moreover, since the first percept does not share any similarities with the third percept, the full wildcard clip ($\#$, $\#$) is created and placed in the second layer, as shown in Fig.~\ref{fig:Fig2}(b) (depicted as a circle with a single $\#$ symbol). Last, at $t=4$, a left-red signal is shown. This causes the creation of the ($\Leftarrow$, \#)-clip (the left-arrow is shared with the first percept) and the (\#, red) clip (the red color is shared with the third clip). After the fourth time step the network is fully established and from this point on will only evolve through changes in the weights of the edges, i.e.\ by modifying the $h$-values.

The mechanism we have described so far, realizes, by construction, the first two characteristics of meaningful generalization: categorization and classification. In particular, categorization, the ability to recognize common properties, is achieved by composing the wildcard clips according to similarities in the coming input. For example, it is natural to think of the ($\#$, red) wildcard clip as representing the common property of \emph{redness}. In that spirit, one could interpret the full wildcard clip ($\#$, $\#$) as representing a general perceptual input. Likewise, classification, the ability to relate a new stimulus to the group of similar past stimuli, is fulfilled there by connecting of the lower-level wildcard clips to matching higher lever wildcard clips, as described above (where percept clips are regarded here as zero-order wildcard clips). Note that classification is done, therefore, not only on the level of the percept clips, but also on the level of the wildcard clips.

While categorization and classification are realized by the very structure of the clip network, the remaining list of requirements, namely, relevant, correct, and flexible generalization, is fulfilled via the update of the $h$-values. To illustrate this on a simple domain, we next confront the agent with four different environmental scenarios, one after the other.  Each scenario lasts 1000 time steps, following by a sudden change of the rewarding scheme, to which the agent has to adapt. The different scenarios are listed below:
\begin{itemize}
    \item[(i)] At the beginning ($1\leq t\leq 1000$), the agent is rewarded for stopping at red light and for driving at green light, irrespective of the arrow direction. 
	 \item[(ii)] At the second phase ($1000<t\leq 2000$), the agent is rewarded for doing the opposite: stopping at green light and driving at red light. 
	\item[(iii)] At the third phase ($2000<t\leq 3000$), the agent should only follow the arrows: it is rewarded for driving (stopping) when the arrow points to the left (right). Colors should thus be ignored. 
	\item[(iv)] In the last phase ($3000<t\leq 4000$), the environment rewards the agent whenever it chooses to drive, irrespective of either the signal's color or its arrow.
\end{itemize}

\begin{figure}[h]
	\centering
	\includegraphics[width=8cm]{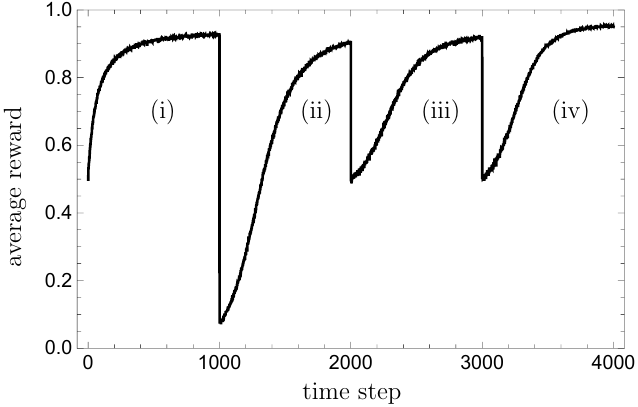}
	\caption{The average reward obtained by the PS agent with generalization as simulated for the four different phases of the driver scenario (see text). At the beginning of each phase, the agent has to adapt to the new rules of the environment. The average reward drops and revives again, thereby exhibiting the mechanism's correctness and flexibility. A damping parameter of $\gamma=0.005$ was used, and the average was taken over $10^4$ agents.}
	\label{fig:Fig4}
\end{figure}

\begin{figure*}[t!]
	\centering
	\includegraphics[width=1\textwidth]{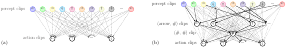}
	\caption{The basic (a) and the enhanced (b) PS networks as they are built up in the neverending-color scenario. (a) Each percept clip at the first row is independently connected to all $n$ action clips at the second row. (b) Each percept- and wildcard-clip is connected to higher-level matching wildcard clips and to all $n$ action clips. For clarity, only one-level edges to and from wildcard clips are solid, while other edges are semitransparent. The thickness of the edges does not reflect their weights.}
	\label{fig:Fig5}
\end{figure*}
 
Figure~\ref{fig:Fig3} sketches four different network configurations that typically develop during the above phases. Only strong edges of relative large $h$-values are depicted, and we ignore direct edges from percepts to actions, for clarity, as explained later. At each stage a different configuration develops, demonstrating how the relevant wildcard clips play an important role, via strong connections to action clips. Moreover, those wildcard clips are connected strongly to the correct action clips. The relevant and correct edges are built through the update rule of Eq.~(\ref{eq:hupdate}), which only strengthens edges that, after having been traversed, lead to a rewarded action. Finally, the presented flexibility in the network's configuration, which reflects a flexible generalization ability, is due to: (a) the existence of all possible wildcard clips in the network; and (b) the update rule of Eq.~(\ref{eq:hupdate}), which allows the network, through a non-zero damping parameter $\gamma$ to adapt fast to changes in the environment. We note that Fig.~\ref{fig:Fig3} only displays idealized network configurations. In practice, other strong edges may exist, e.g.\ direct edges from percepts to actions, which may be rewarded as well. In the next section we address such alternative configurations and analyze their influence on the agent's success rate.

Figure~\ref{fig:Fig4} shows the PS agent's performance, that is, the average reward obtained by the agent, in the driver scenario, as a function of time, averaged over $10^4$ agents. A reward of $\lambda=1$ is given for correct actions and a damping parameter of $\gamma = 0.005$ is used. In the PS model there is a trade off between adaptation time and the maximally achievable average reward. A high damping parameter, $\gamma$, leads to faster relearning, but to lower averaged asymptotic success rates, see also Ref.~\cite{mautner2013projective}. Here we chose $\gamma = 0.005$ to allow the network to adapt within $1000$ time steps. It is shown that on average the agent managed to adapt to each of the phases imposed by the environment, and to learn the correct actions. We can also see that the asymptotic performance of the agent is slightly better in the last phase, where the correct action is independent from the input. To understand this, note that: (a) The relevant edge can be rewarded at each time step and thus be less affected by the non-zero damping parameter; and (b) Each wildcard clip necessarily leads to the correct action. This observation indicates that the agent's performance improves when the stimuli can be generalized to a greater extent. We will encounter this feature once more in the next section, where it is analytically verified.

The driver's scenario is given here to explain and demonstrate the underlying principles of the proposed generalization mechanism within PS. The problem itself can, of course, be solved by the basic PS alone, as well as by other methods, without a generalization capability. In what follows, however, we consider scenarios in which such an ability to generalize is indeed crucial for the agent's success.

\subsection*{Experimental and analytic results}

\subsubsection*{A simple example}

Sometimes it is not only helpful but a plain necessity for the agent to have a mechanism of generalization, as otherwise it has no chance to learn. Consider, for example, a situation in which the agent perceives a new stimulus every time step. What option does it have, other than trying to find some similarities among those stimuli, upon which it can act?  In this section we consider such a scenario and analyze it in detail. Specifically, the environment presents one of $n$ different arrows, but at each time step the background color is different. The agent can only move into one of the $n>1$ corresponding directions and the environment rewards the agent whenever it follows the direction of an arrow, irrespective of its color. We call it the neverending-color scenario. 

In this scenario, the basic PS agent has a two-layered clip network, of the structure presented in Fig.~\ref{fig:Fig5}(a). At each time step, a new percept clip is created, from which the random walk leads, after a single transition, to one of the $n$ possible action clips the agent can perform. The problem is that even if the agent takes the correct direction, the rewarded edge will never take part in later time steps, as no symbol is shown twice. The basic PS agent has thus no other option but to choose an action at random, which will be correct only with probability of $1/n$, even after infinitely many time steps. In contrast to the basic PS, the PS with generalization does show learning behavior. The full network is shown in Fig.~\ref{fig:Fig5}(b). Percept clips and wildcard clips are connected to matching wildcard clips and to all actions. Note that the wildcard clips ($\#$, color) are never created, as no color is seen twice. 

\begin{figure*}[t!]
	\centering
	\includegraphics[height=3.1cm]{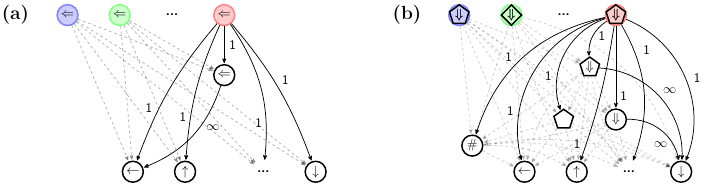}
	\caption{(a) The enhanced PS network as it is built up for the neverending-color scenario with $K=2$ categories. Only the subnetwork corresponding to the left-arrow is shown. The weight of the edge from the wildcard clip ($\Leftarrow$, $\#$) to the correct action clip ($\leftarrow$) goes to $h=\infty$ with time. Hopping to the ($\Leftarrow$, $\#$) clip then leads to the rewarded action with certainty. Otherwise, hopping randomly to any of the other clips is only successful with probability $1/n$. 
	(b) The enhanced PS network as it is built up for the neverending-color scenario with $K=3$ categories (direction, color, shape). Only the subnetwork corresponding to the down-arrow is shown. The weights of the edges from the wildcard clips ($\Downarrow$, $\#$, pentagon) and ($\Downarrow$, $\#$, $\#$), in the first and second layer, respectively, to the correct action clip ($\downarrow$), go to $h=\infty$ with time. 
	(a) and (b): Edges that are relevant for the analysis are solid, whereas other edges are semitransparent. The thickness of the edges does not reflect their weights.}
	\label{fig:Fig6}
\end{figure*}

To illustrate the fundamental difference in performance between the basic PS and the PS with generalization we consider their asymptotic efficiencies. As explained above, the basic PS agent can only be successful with probability $1/n$. To see that the PS agent with generalization can do better, we take a closer look on the (arrow, $\#$) clips. These clips will, eventually, have very strong edges to the correct action clip. In fact, in the case of zero damping ($\gamma=0$) we consider here, the $h$-values of these edges will tend to infinity with time, implying that once an (arrow, $\#$) clip is hit, the probability to hop to the correct action clip becomes unity. This is illustrated for the left-arrow case in Fig.~\ref{fig:Fig6}(a). 

At each time step, the agent is confronted with a certain colored arrow. The corresponding new percept clip is created and a random walk on the network begins. To determine the exact asymptotic performance of the PS agent with generalization, we should consider two possibilities: Either the wildcard corresponding clip (arrow, $\#$) is hit, or it is not. In the first case, which occurs with probability $p=1/(n+2)$, the excitation will hop to the correct action with unit probability and the agent will be rewarded. In the second case, no action is preferred over the others and the correct action will be reached with probability  $1/n$. It is possible that an edge from the full wildcard clip ($\#$, $\#$) to some action clip was previously rewarded, yet when averaging over all agents we still get an averaged success probability of $1/n$. Overall, the performance of the PS agent with generalization is thus given by:
\begin{equation}
 \mathcal{E}_\infty(n) = p + \left(1-p\right)\frac{1}{n} = \frac{1+2n}{n(n+2)} > \frac{1}{n},~~~p=\frac{1}{n+2}, 
 \label{eq:Asymptotics2cat}
\end{equation}
which is independent of the precise value of the reward $\lambda$ (as long as it is a positive constant). Recall that $n>1$ in the considered environment.

\begin{figure*}[t!]
	\centering
	\includegraphics[width=17.7cm]{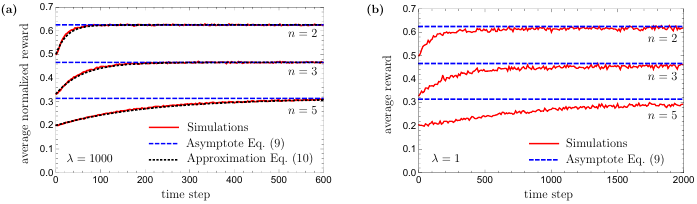}
	\caption{Learning curves of the PS agent with generalization in the neverending-color scenario for $n=2, 3$ and $5$ actions. Simulations of $10^5$ agents are shown in red, asymptotic average reward lines $\mathcal{E}_\infty(n)$ (Eq.~(\ref{eq:Asymptotics2cat})) are shown in dashed blue, the corresponding analytical approximation curves (Eq.~(\ref{eq:AnalyticalApproximation})) are shown in dotted black. (a) A high reward of $\lambda=1000$ was used. (b) A reward of $\lambda=1$ was used.}
	\label{fig:Fig7}
\end{figure*}

In Fig.~\ref{fig:Fig7}(a), the average performance of the PS agent with generalization, obtained through numerical simulation, is plotted as a function of time, in solid red curves, for several values of $n$. Initially, the average performance is $1/n$, i.e.\ completely random (which is the best performance of the basic agent). It then grows, indicating that the agent begin to learn how to respond correctly, until it reaches its asymptotic value, as given in Eq.~(\ref{eq:Asymptotics2cat}) and marked in the figure with a dashed blue line. It is seen that in these cases, the asymptotic performance is achieved already after tens to hundreds of time steps (see the next section for an  analytical expression of the learning rate). The simulations were carried out with $10^5$ agents and a zero damping parameter ($\gamma=0$). Since the asymptotic performance of Eq.~(\ref{eq:Asymptotics2cat}) is independent of the reward $\lambda$ and to ease the following analytical analysis, we chose a high value of $\lambda=1000$. Setting a smaller reward would only amount to a slower learning curve, but with no qualitative difference, as one can see in Fig.~\ref{fig:Fig7}(b) for a reward of $\lambda=1$. In Fig.~\ref{fig:Fig7}(a) and whenever the reward $\lambda$ is not $1$, we normalize the average reward obtained by the PS agent in order to compare to the probability of obtaining the maximum reward given by Eq.~(\ref{eq:Asymptotics2cat}). 
 
\begin{figure*}[t!]
	\centering
	\includegraphics[width=17.7cm]{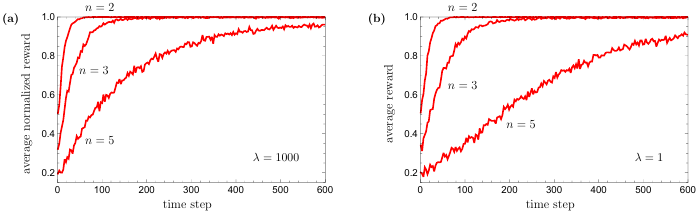}
	\caption{Learning curves of the PS agent with generalization augmented with majority voting in the neverending-color scenario for $n=2, 3$ and $5$ actions. Curves are obtained by averaging over $10^3$ agents, majority voting consists of $100$ random walks. In order to achieve shorter learning times, the creation of the (\#, \#)-clip was suppressed in these simulation. (a) A high reward of $\lambda=1000$ was used. (b) A reward of $\lambda=1$ was used.}
	\label{fig:Fig8}
\end{figure*}

We have therefore shown that the generalization mechanism leads to a clear qualitative advantage in this scenario: without it the agent can not learn, whereas with it, it can. As for the quantitative difference, both Eq.~(\ref{eq:Asymptotics2cat}) and Fig.~\ref{fig:Fig7}(a) indicate that the gap in performance is not large. Nonetheless, any such gap can be further amplified. The idea is that for any network configuration of which the probability to take the correct action is larger than the probability to take any other action, the success probability can be amplified to unity by ``majority voting", i.e.\ by performing the random walk several times, and choosing the action clip that occurs most frequently. Such amplification rapidly increases the agent's performance whenever the gap over the fully random agent is not negligible given that the full wildcard clip ($\#$, $\#$) is not used. In Fig.~\ref{fig:Fig8} we show the simulations of the PS agent with majority voting. The results demonstrate the convergence to the optimal performance, regardless of the value of the reward.

The presented performance of the PS model (including generalization, but without any external classifying machinery) in the neverending-color scenario can in principle be contrasted to other standard ``raw"  RL models. However this is clearly not a reasonable comparison, because standard RL models, e.g. SARSA or Q-learning, cannot learn anything in this task environment without an external classifying machinery (decision trees, neural networks or SVM). On the flip side, the capacity of a combination of basic RL models with classifier machinery may beat the performance of our model. But again, this is arguably not an instructive comparison as we do not combine the PS with external classifier machinery. The latter could in principle be done, but all such solutions come at the price of loss of homogeneity, and (sometimes dramatic) increase in model complexity, which we avoid in this work for the reasons we elaborated on earlier.

\subsubsection*{Analytical analysis: learning curve}

To analyze the PS learning curves and to predict the agent's behavior for arbitrary $n$, we take the following simplifying assumptions: First, we assume that all possible wildcard clips are present in the PS network from the very beginning; second, for technical reasons, we assume that edges from the partial wildcard clips (arrow, $\#$) to the full wildcard clip ($\#$, $\#$) are never rewarded; last, we set an infinite reward $\lambda = \infty$ (or, equivalently, one can use the softmax function in Eq.~(\ref{eq:probablititesSoftmax}) with $\beta = \infty$ instead of Eq.~(\ref{eq:probablititesBasic})). As shown below, under these assumptions, the analysis results in a good approximation of the actual performance of the agent. 

While Eq.~(\ref{eq:Asymptotics2cat}) provides an expression for $\mathcal{E}_{\infty}(n)$, the expected average reward of the agent at infinity, here we look for the expected average reward at any time step $t$, i.e.\ we look for $\mathcal{E}_{t}(n)$. 
Taking the above assumptions into account and following the same arguments that led to Eq.~(\ref{eq:Asymptotics2cat}), we note that at any time $t$, at which the arrow $a$ is shown, there are only two possible network configurations that are conceptually different: Either the edge from the ($a$, $\#$) clip to the ($a$) action clip was already rewarded and has an infinite weight, or not.  Note that while this edge must eventually be rewarded, at any finite $t$ this is not promised.  
Let $p_\mathrm{learn}(t)$ be the probability that the edge from the wildcard clip ($a$, \#) to the action clip ($a$) has an infinite $h$-value at time $t$, i.e.\ the probability that the correct association was learned, then the expected performance at time $t$ is given by:
\begin{equation}
\mathcal{E}_t(n) = p_\mathrm{learn}(t)~\mathcal{E}_\infty(n) + \left(1-p_\mathrm{learn}(t)\right)\frac{1}{n}.
 \label{eq:AnalyticalApproximation}
\end{equation}
The probability $p_\mathrm{learn}(t)$ can be written as
\begin{equation}
p_\mathrm{learn}(t) = 1- \left(1-\frac{1}{n(n+1)(n+2)}\right)^{t-1},
 \label{eq:AnalyticalApproximation2}
\end{equation}
where the term $1/n(n+1)(n+2)$ corresponds to the probability of finding the rewarded path (labeled as ``$\infty$'' in Fig.~\ref{fig:Fig6}(a)): $1/n$ is the probability that the environment presents the arrow ($x$), then the probability to hop from the percept clip ($x$, color) to the wildcard clip  ($x$, $\#$) is  $1/(n+2)$, and the probability to hop from the wildcard clip ($x$, $\#$) to the ($x$) action clip is $1/(n+1)$. Finally, we take into account the fact that, before time $t$, the agent had $(t-1)$ opportunities to take this path and be rewarded. 

The analytical approximation of the time-dependent performance of PS, given in Eq.~(\ref{eq:AnalyticalApproximation}) is plotted on top of Fig.~\ref{fig:Fig7}(a) in dotted black, where it is shown to match the simulated curves well (in red). The difference, which one can see in detail in Fig.~\ref{fig:Fig9}(a), in the very beginning is caused by the assumption that all wildcard clips are present in the network from the very beginning, whereas the real agent needs several time steps to create them, thus reducing its initial success probability. Nonetheless, after a certain number of time-steps the simulated PS agent starts outperforming the prediction given by the analytic approximation, because the agent can be rewarded for transition from the wildcard clip (arrow, \#) to the full wildcard clip (\#, \#), leading to higher success probability. Note that despite the described difference for early time steps, which is shown in Fig.~\ref{fig:Fig9}(a), both curves converge to the same value as one can see in Fig.~\ref{fig:Fig7}(a).

\subsubsection*{Analytical analysis: learning time}

The best performance a single PS agent can achieve is given by the asymptotic performance $\mathcal{E}_\infty(n)$ (Eq.~(\ref{eq:Asymptotics2cat})). For each agent there is a certain finite time $\tau$ at which this performance is achieved, thus all agents reach this performance at $t=\infty$. $p_\mathrm{learn}(t)$, as defined before, is the probability that the edge from the relevant wildcard clip to the correct action clip was rewarded up to time $t$, and can be expressed as a cumulative distribution function $P(\tau \leq t-1)$, so that 
$P(\tau=t)=P(\tau \leq t) - P(\tau \leq t-1) = p_\mathrm{learn}(t+1) - p_\mathrm{learn}(t)$.

The expected value of $\tau$ can be thought of as the \emph{learning time} of the agent and can be expressed as a power series
\begin{eqnarray}
& \mathrm{E}[\tau] = \sum_{t=1}^\infty tP(\tau=t) 
= \sum_{t=1}^\infty t\Biggl(\left(1-\frac{1}{n(n+1)(n+2)}\right)^{t-1}\nonumber \\
& - \left(1-\frac{1}{n(n+1)(n+2)}\right)^{t} \Biggr) = n(n+1)(n+2).
\label{eq:expectedValue}
\end{eqnarray}
Note that the ``not learning" probability ($1-p_{\mathrm{learn}}(t)$) decays exponentially, with the decay rate reflected in the learning time $\mathrm{E}[\tau]$ (see Eq.~(\ref{eq:AnalyticalApproximation2})).

\subsubsection*{More than two categories} 

We next generalize the neverending-color task to the case of an arbitrary number of categories $K$. The color category may take infinite values, whereas any other category can only take finite values, and the number of possible actions is given by $n>1$. As before, only one category is important, namely the arrow direction, and the agent is rewarded for following it, irrespective of all other input. With more irrelevant categories the environment thus overloads the agent with more unnecessary information, would this affect the agent's performance?

To answer this question, we look for the corresponding averaged asymptotic performance. As before, in the limit of $t\to\infty$, the wildcard clips which contain the arrows lead to a correct action clip with unit probability (for any finite reward $\lambda$ and zero damping parameter $\gamma=0$), as illustrated in Fig.~\ref{fig:Fig6}(b). On the other hand, choosing any other clip (including action clips) results with the correct action  with an averaged probability of only $1/n$. Accordingly, either the random walk led to a wildcard clip with an arrow, or it did not. The averaged asymptotic performance for $K$ categories and $n$ actions can hence be written as
 \begin{equation}
 \mathcal{E}_\infty(n,K) =  p + \left(1-p\right)\frac{1}{n} = \frac{n+(1+n)2^{K-2}}{n\left(n+2^{K-1}\right)}, 
 \label{eq:AsymptoticsKcat}
\end{equation}
where $p=2^{K-2}/(n+2^{K-1})$ is the probability to hit a wildcard clip with an arrow, given by the ratio between the number of wildcard clips with an arrow ($2^{K-2}$), and the total number of clips that are reachable from a percept clip ($n+2^{K-1}$). In this scenario, where no color is shown twice, all wildcard clips have their color category fixed to a wildcard symbol $\#$. There are thus $2^{K-1}$ wildcard clips that are connected to each  percept clip, where half of them, i.e.\ $2^{K-2}$ contain an arrow. Note that for two categories Eq.~(\ref{eq:AsymptoticsKcat}) correctly reduces to Eq.~(\ref{eq:Asymptotics2cat}).

\begin{figure*}[t!]
	\centering
	\includegraphics[width=17.7cm]{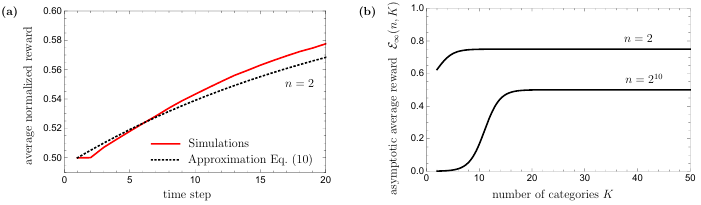}
	\caption{(a) A part of Fig.~\ref{fig:Fig7}(a): The difference between the PS agent's performance approximation, which is given by Eq.~(\ref{eq:AnalyticalApproximation}), and numerical simulations for $n=2$. (b) The asymptotic average reward $\mathcal{E}_\infty (n, K)$ for the neverending-color scenario (see Eq.~\ref{eq:AsymptoticsKcat}), as a function of $K$, the number of categories, for $n=2, 2^{10}$.}
	\label{fig:Fig9}
\end{figure*}

We can now see the effect of having a large number $K$ of irrelevant categories on the asymptotic performance of the agent. First, it is easy to show that for a fixed $n$, $\mathcal{E}_\infty(n,K)$ increases monotonically with $K$, as also illustrated in Fig.~\ref{fig:Fig9}(b). This means that although the categories provided by the environment are irrelevant, the generalization machinery can exploit them to make a larger number of relevant generalizations, and thereby increase the agent's performance. Moreover, for large $K$, and, more explicitly, for $K\gg\log n$, the averaged asymptotic performance tends to $\left(1+1/n\right)/2$. Consequently, when the number of possible actions $n$ is also large, in which case the performance of the basic agent would drop to $0$, the PS agent with generalization would succeed with a probability that tends to $1/2$, as shown in Fig.~\ref{fig:Fig9}(b) for $n=2^{10}$.

Similarly, we note that when none of the categories is relevant, i.e.\ when the environment is such that the agent is expected to take the same action irrespective of the stimulus it receives, the agent performs even better, with an average asymptotic performance of $\mathcal{E}^*_\infty(n,K) = \left(1+2^{K-1}\right)/\left(n+2^{K-1}\right)$. This is because in such a scenario, every wildcard clip eventually connects to the rewarded action. Accordingly, since each percept clip leads to wildcard clips with high probability, the correct action clip is likely to be reached. In fact, in the case of $K\gg\log n$ the asymptotic performance of the PS with generalization actually tends to $1$.

\section*{Discussion}

When the environment confronts an agent with a new stimulus in each and every time step, the agent has no chance of coping, unless the presented stimuli have some common features that the agent can grasp. The recognition of these common features, i.e.\ categorization, and classifying new stimuli accordingly, are the first steps toward a meaningful generalization as characterized at the beginning of this paper. We presented a simple dynamical machinery that enables the PS model to realize those abilities and showed how the latter requirements of meaningful generalization - that relevant generalizations are learned, that correct actions are associated with the relevant properties, and that the generalization mechanism is flexible - follow naturally from the PS model itself. Through numerical and analytical analysis, also for an arbitrary number of categories, we showed that the PS agent can then learn even in extreme scenarios where each percept is presented only once. 

The generalization machinery introduced in this paper, enriches the basic PS model: not only in the practical sense, i.e.\ that it can handle a larger range of scenarios, but also in a more conceptual sense. 
In particular, the enhanced PS network allows for the emergence of clips that represent \emph{abstractions} or \emph{abstract properties}, like the redness property, rather then merely remembered percepts or actions. Moreover, the enhanced PS network is multilayered and allows for more involved dynamics of the random walk, which, as we have shown, gives rise to a more sophisticated behavior of the agent.
Yet, although the clip network may evolve to more complicated structures than before, the overall model preserves its inherent simplicity, which enables an analytical characterization of its  performance. 

Our approach assumes that the percept space has an underlying Cartesian product structure, that is, there are established categories. This is a natural assumption in many settings, for instance where the categories stem from distinct sensory devices, in the case of embodied agents.
For the generalization procedure to produce functional generalizations, however, it is also vital that the reward function of the environment indeed does reflect this structure. This implies that the status of some categories matters, where other categories may not matter. Nonetheless, in the settings where the underlying similarity structure on the percept space (e.g. a metric which establishes which subsets of the percept space require similar responses) does not follow the Cartesian feature axes, the agent will still learn, provided the percept space is not infinite. In this case, no improvements will come about from the generalization procedure. While restricted, the provided notion of generalization still captures an exponentially large collection of subsets of the percept space, in the number of categories.

Regarding the computational complexity of the model, we can identify two main contributions. The first is the deliberation time of the model -- that is, the number of transitions within the ECM which have to occur at any time-step of the interactions. This is efficient, as the number of steps is upper bounded by the number of layers, which scales linearly with the number of categories.
The second contribution stems from the updates of the network -- the number of wildcard clips which have to be generated -- which also has an immediate impact on the expected learning time of the agent. In principle, the total number of wildcard clips can grow exponentially with the number of categories (that is, typically as a low order polynomial in the size of the percept space). Note that this is a necessity, as there are exponentially many subspaces of the percept space which the agent must explore to find the relevant generalizations. Confining this space to a smaller subspace would necessarily restrict the generality of the generalization procedure. Thus there is an unavoidable trade-off:  how general is the generalization procedure versus how large is the generalization space -- one has to balance.

In the approach to generalization we have presented, we have built upon the capacity of the PS model itself to dynamically generate novel clips, and the restriction we have mentioned (combinatorial space complexity due to the overall clip number) could be mitigated by employing external classification machinery. This latter approach is the norm in other RL approaches, but as we have clarified earlier, it comes with other types of issues we have avoided: the use of external machinery increases the complexity of the model, and makes it inhomogeneous. In contrast, the homogeneity of the PS approach allows for high interpretability of results, including analytic analyses, and advantages with respect to embodied implementations. Moreover, our approach has a clear route for quantization of the model, which may become increasingly more relevant as quantum technologies further develop. Thus, it is of great interest to develop a method that can deal with the issue of combinatorial space complexity while maintaining homogeneity.

In particular, a clip-deletion mechanism can be used for PS to deal with the combinatorial growth of the number of clips. This mechanism deletes clips, thereby maintaining a stable population of a controlled size and prioritizing the deletion of less used, hence less useful clips. The size of the population constitutes a sparsity parameter of the model, and makes the combinatorial explosion in the clip number controllable, while still allowing the agent to explore the complete space of axes-specified generalizations. A proof of principle of this approach was given in Ref.~\cite{simon2015evaluation}; a more detailed analysis of the deletion mechanism is ongoing work.

\section*{Acknowledgements}

We wish to thank Markus Tiersch, Dan Browne and Elham Kashefi for helpful discussions. 
This work was supported in part by the Austrian Science Fund (FWF) through Grant No. SFB FoQuS F4012, and by the Templeton World Charity Foundation (TWCF) through Grant No. TWCF0078/AB46.


\providecommand{\noopsort}[1]{}\providecommand{\singleletter}[1]{#1}


\begin{thebibliography}{10}
\expandafter\ifx\csname url\endcsname\relax
  \def\url#1{\texttt{#1}}\fi
\expandafter\ifx\csname urlprefix\endcsname\relax\def\urlprefix{URL }\fi
\providecommand{\bibinfo}[2]{#2}
\providecommand{\eprint}[2][]{\url{#2}}

\bibitem{Holland_etal_Induction86}
\bibinfo{author}{Holland, J.~H.}, \bibinfo{author}{Holyoak, K.~J.},
  \bibinfo{author}{Nisbett, R.~E.} \& \bibinfo{author}{Thagard, P.}
\newblock \emph{\bibinfo{title}{Induction: Processes of Inference, Learning,
  and Discovery}}.
\newblock Computational Models of Cognition and Perception
  (\bibinfo{publisher}{MIT Press}, \bibinfo{address}{Cambridge, MA, USA},
  \bibinfo{year}{1986}).

\bibitem{2013_Saitta_book}
\bibinfo{author}{Saitta, L.} \& \bibinfo{author}{Zucker, J.-D.}
\newblock \emph{\bibinfo{title}{Abstraction in Artificial Intelligence and
  Complex Systems}} (\bibinfo{publisher}{Springer}, \bibinfo{address}{New York,
  NY, USA}, \bibinfo{year}{2013}).

\bibitem{barto1998reinforcement}
\bibinfo{author}{Sutton, R.~S.} \& \bibinfo{author}{Barto, A.~G.}
\newblock \emph{\bibinfo{title}{Reinforcement Learning: An Introduction}}
  (\bibinfo{publisher}{MIT press}, \bibinfo{address}{Cambridge, MA, USA},
  \bibinfo{year}{1998}).

\bibitem{russell2010artificial}
\bibinfo{author}{Russell, S.} \& \bibinfo{author}{Norvig, P.}
\newblock \emph{\bibinfo{title}{Artificial Intelligence: A Modern Approach}}
  (\bibinfo{publisher}{Prentice Hall}, \bibinfo{address}{Englewood Cliffs, NJ,
  USA}, \bibinfo{year}{2010}), \bibinfo{edition}{third} edn.

\bibitem{RL_2012_book}
\bibinfo{editor}{Wiering, M.} \& \bibinfo{editor}{van Otterlo, M.} (eds.)
  \emph{\bibinfo{title}{Reinforcement Learning: State-of-the-Art}},
  vol.~\bibinfo{volume}{12} of \emph{\bibinfo{series}{Adaptation, Learning, and
  Optimization}} (\bibinfo{publisher}{Springer}, \bibinfo{address}{Berlin,
  Germany}, \bibinfo{year}{2012}).

\bibitem{2008_van_Otterlo}
\bibinfo{author}{van {Otterlo}, M.}
\newblock \emph{\bibinfo{title}{The logic of adaptive behavior: knowledge
  representation and algorithms for the Markov decision process framework in
  first-order domains}}.
\newblock Ph.D. thesis, \bibinfo{school}{Univ. Twente},
  \bibinfo{address}{Enschede, Netherlands} (\bibinfo{year}{2008}).

\bibitem{2010_Ponsen}
\bibinfo{author}{Ponsen, M.}, \bibinfo{author}{Taylor, M.~E.} \&
  \bibinfo{author}{Tuyls, K.}
\newblock \bibinfo{title}{Abstraction and generalization in reinforcement
  learning: A summary and framework}.
\newblock In \bibinfo{editor}{Taylor, M.~E.} \& \bibinfo{editor}{Tuyls, K.}
  (eds.) \emph{\bibinfo{booktitle}{Adaptive and Learning Agents}}, vol.
  \bibinfo{volume}{5924} of \emph{\bibinfo{series}{Lecture Notes in Computer
  Science}}, chap.~\bibinfo{chapter}{1}, \bibinfo{pages}{1--32}
  (\bibinfo{publisher}{Springer}, \bibinfo{address}{Berlin, Germany},
  \bibinfo{year}{2010}).

\bibitem{Watkins_1989}
\bibinfo{author}{Watkins, C. J. C.~H.}
\newblock \emph{\bibinfo{title}{Learning from delayed rewards}}.
\newblock Ph.D. thesis, \bibinfo{school}{Univ. Cambridge},
  \bibinfo{address}{Cambridge, U.K.} (\bibinfo{year}{1989}).

\bibitem{rummery1994line}
\bibinfo{author}{Rummery, G.~A.} \& \bibinfo{author}{Niranjan, M.}
\newblock \bibinfo{title}{On-line {Q}-learning using connectionist systems}.
\newblock \bibinfo{type}{Tech. Rep.} \bibinfo{number}{CUED/F-INFENG/TR 166},
  \bibinfo{institution}{Univ. Cambridge}, \bibinfo{address}{Cambridge, U.K.}
  (\bibinfo{year}{1994}).

\bibitem{melo2008analysis}
\bibinfo{author}{Melo, F.~S.}, \bibinfo{author}{Meyn, S.~P.} \&
  \bibinfo{author}{Ribeiro, M.~I.}
\newblock \bibinfo{title}{An analysis of reinforcement learning with function
  approximation}.
\newblock In \emph{\bibinfo{booktitle}{Proc. 25th Int. Conf. Mach. Learn.}},
  \bibinfo{pages}{664--671} (\bibinfo{year}{2008}).

\bibitem{1975_Albus_CMAC}
\bibinfo{author}{Albus, J.~S.}
\newblock \bibinfo{title}{A new approach to manipulator control: The cerebellar
  model articulation controller \uppercase{(CMAC)}}.
\newblock \emph{\bibinfo{journal}{J. Dyn. Sys., Meas., Control.}}
  \textbf{\bibinfo{volume}{97}}, \bibinfo{pages}{220--227}
  (\bibinfo{year}{1975}).

\bibitem{sutton1996generalization}
\bibinfo{author}{Sutton, R.~S.}
\newblock \bibinfo{title}{Generalization in reinforcement learning: Successful
  examples using sparse coarse coding}.
\newblock In \emph{\bibinfo{booktitle}{Adv. Neural Inf. Process. Syst.}},
  vol.~\bibinfo{volume}{8}, \bibinfo{pages}{1038--1044}
  (\bibinfo{publisher}{MIT Press}, \bibinfo{year}{1996}).

\bibitem{boyan1995generalization}
\bibinfo{author}{Boyan, J.~A.} \& \bibinfo{author}{Moore, A.~W.}
\newblock \bibinfo{title}{Generalization in reinforcement learning: Safely
  approximating the value function}.
\newblock In \emph{\bibinfo{booktitle}{Adv. Neural Inf. Process. Syst.}},
  vol.~\bibinfo{volume}{7}, \bibinfo{pages}{369--376} (\bibinfo{publisher}{MIT
  Press}, \bibinfo{year}{1995}).

\bibitem{whiteson2006evolutionary}
\bibinfo{author}{Whiteson, S.} \& \bibinfo{author}{Stone, P.}
\newblock \bibinfo{title}{Evolutionary function approximation for reinforcement
  learning}.
\newblock \emph{\bibinfo{journal}{J. Mach. Learn. Res.}}
  \textbf{\bibinfo{volume}{7}}, \bibinfo{pages}{877--917}
  (\bibinfo{year}{2006}).

\bibitem{2015_QL_Nature}
\bibinfo{author}{Mnih, V.} \emph{et~al.}
\newblock \bibinfo{title}{Human-level control through deep reinforcement
  learning}.
\newblock \emph{\bibinfo{journal}{Nature}} \textbf{\bibinfo{volume}{518}},
  \bibinfo{pages}{529--533} (\bibinfo{year}{2015}).

\bibitem{Pyeatt98decisiontree}
\bibinfo{author}{Pyeatt, L.~D.} \& \bibinfo{author}{Howe, A.~E.}
\newblock \bibinfo{title}{Decision tree function approximation in reinforcement
  learning}.
\newblock In \emph{\bibinfo{booktitle}{Proc. 3rd Int. Symposium on Adaptive
  Systems: Evolutionary Computation and Probabilistic Graphical Models}},
  \bibinfo{pages}{70--77} (\bibinfo{year}{2001}).

\bibitem{2005_Q_iteration}
\bibinfo{author}{Ernst, D.}, \bibinfo{author}{Geurts, P.} \&
  \bibinfo{author}{Wehenkel, L.}
\newblock \bibinfo{title}{Tree-based batch mode reinforcement learning}.
\newblock \emph{\bibinfo{journal}{J. Mach. Learn. Res.}}
  \textbf{\bibinfo{volume}{6}}, \bibinfo{pages}{503--556}
  (\bibinfo{year}{2005}).

\bibitem{Utgoff98constructive}
\bibinfo{author}{Utgoff, P.~E.} \& \bibinfo{author}{Precup, D.}
\newblock \bibinfo{title}{Constructive function approximation}.
\newblock In \bibinfo{editor}{Liu, H.} \& \bibinfo{editor}{Motoda, H.} (eds.)
  \emph{\bibinfo{booktitle}{Feature Extraction, Construction and Selection}},
  vol. \bibinfo{volume}{453} of \emph{\bibinfo{series}{The Springer
  International Series in Engineering and Computer Science}},
  \bibinfo{pages}{219--235} (\bibinfo{publisher}{Springer},
  \bibinfo{address}{New York, NY, USA}, \bibinfo{year}{1998}).

\bibitem{1995_Vapnik}
\bibinfo{author}{Cortes, C.} \& \bibinfo{author}{Vapnik, V.}
\newblock \bibinfo{title}{Support-vector networks}.
\newblock \emph{\bibinfo{journal}{Mach. Learn.}} \textbf{\bibinfo{volume}{20}},
  \bibinfo{pages}{273--297} (\bibinfo{year}{1995}).

\bibitem{2007_Laumonier}
\bibinfo{author}{Laumonier, J.}
\newblock \bibinfo{title}{Reinforcement using supervised learning for policy
  generalization}.
\newblock In \emph{\bibinfo{booktitle}{Proc. 22nd National Conference on
  Artificial Intelligence}}, vol.~\bibinfo{volume}{2},
  \bibinfo{pages}{1882--1883} (\bibinfo{publisher}{AAAI Press},
  \bibinfo{year}{2007}).

\bibitem{Holland76}
\bibinfo{author}{Holland, J.~H.}
\newblock \bibinfo{title}{Adaptation}.
\newblock In \bibinfo{editor}{Rosen, R.~J.} \& \bibinfo{editor}{Snell, F.~M.}
  (eds.) \emph{\bibinfo{booktitle}{Progress in Theoretical Biology}},
  vol.~\bibinfo{volume}{4}, \bibinfo{pages}{263--293} (\bibinfo{year}{1976}).

\bibitem{Holland86}
\bibinfo{author}{Holland, J.~H.}
\newblock \bibinfo{title}{Escaping brittleness: The possibilities of
  general-purpose learning algorithms applied to parallel rule-based systems}.
\newblock In \bibinfo{editor}{Michalski, R.~S.}, \bibinfo{editor}{Carbonell,
  J.~G.} \& \bibinfo{editor}{Mitchell, T.~M.} (eds.)
  \emph{\bibinfo{booktitle}{Machine Learning: An Artificial Intelligence
  Approach}}, vol.~\bibinfo{volume}{2} (\bibinfo{publisher}{Morgan Kaufmann},
  \bibinfo{year}{1986}).

\bibitem{2009_LCS_Review}
\bibinfo{author}{Urbanowicz, R.~J.} \& \bibinfo{author}{Moore, J.~H.}
\newblock \bibinfo{title}{Learning classifier systems: A complete introduction,
  review, and roadmap}.
\newblock \emph{\bibinfo{journal}{Journal of Artificial Evolution and
  Applications}} \textbf{\bibinfo{volume}{2009}}, \bibinfo{pages}{1--25}
  (\bibinfo{year}{2009}).

\bibitem{2005_State_abstraction}
\bibinfo{author}{Jong, N.~K.}
\newblock \bibinfo{title}{State abstraction discovery from irrelevant state
  variables}.
\newblock In \emph{\bibinfo{booktitle}{Proc. 19th International Joint
  Conference on Artificial Intelligence}}, \bibinfo{pages}{752--757}
  (\bibinfo{year}{2005}).

\bibitem{2006_State_abstraction}
\bibinfo{author}{Li, L.}, \bibinfo{author}{Walsh, T.~J.} \&
  \bibinfo{author}{Littman, M.~L.}
\newblock \bibinfo{title}{Towards a unified theory of state abstraction for
  \uppercase{MDP}s}.
\newblock In \emph{\bibinfo{booktitle}{Proc. 9th International Symposium on
  Artificial Intelligence and Mathematics}}, \bibinfo{pages}{531--539}
  (\bibinfo{year}{2006}).

\bibitem{2011_State_abstraction}
\bibinfo{author}{Cobo, L.~C.}, \bibinfo{author}{Zang, P.},
  \bibinfo{author}{Isbell, C.~L.} \& \bibinfo{author}{Thomaz, A.~L.}
\newblock \bibinfo{title}{Automatic state abstraction from demonstration}.
\newblock In \emph{\bibinfo{booktitle}{Proc. 22nd International Joint
  Conference on Artificial Intelligence}} (\bibinfo{year}{2011}).

\bibitem{Sutton1999181}
\bibinfo{author}{Sutton, R.~S.}, \bibinfo{author}{Precup, D.} \&
  \bibinfo{author}{Singh, S.}
\newblock \bibinfo{title}{Between \uppercase{MDP}s and semi-\uppercase{MDP}s: A
  framework for temporal abstraction in reinforcement learning}.
\newblock \emph{\bibinfo{journal}{Artif. Intell.}}
  \textbf{\bibinfo{volume}{112}}, \bibinfo{pages}{181 -- 211}
  (\bibinfo{year}{1999}).

\bibitem{2012_HRL}
\bibinfo{author}{Botvinick, M.~M.}
\newblock \bibinfo{title}{Hierarchical reinforcement learning and decision
  making}.
\newblock \emph{\bibinfo{journal}{Curr. Opin. Neurobiol.}}
  \textbf{\bibinfo{volume}{22}}, \bibinfo{pages}{956 -- 962}
  (\bibinfo{year}{2012}).

\bibitem{2004_Tadepalli_relationalRL}
\bibinfo{author}{Tadepalli, P.}, \bibinfo{author}{Givan, R.} \&
  \bibinfo{author}{Driessens, K.}
\newblock \bibinfo{title}{Relational reinforcement learning: An overview}.
\newblock In \emph{\bibinfo{booktitle}{Proc. Int. Conf. Mach. Learn. Workshop
  on Relational Reinforcement Learning}} (\bibinfo{year}{2004}).

\bibitem{hutter2009feature}
\bibinfo{author}{Hutter, M.}
\newblock \bibinfo{title}{Feature reinforcement learning: Part \uppercase{I}.
  \uppercase{U}nstructured \uppercase{MDP}s}.
\newblock \emph{\bibinfo{journal}{Journal of Artificial General Intelligence}}
  \textbf{\bibinfo{volume}{1}}, \bibinfo{pages}{3--24} (\bibinfo{year}{2009}).

\bibitem{2012_Nguyen}
\bibinfo{author}{Nguyen, P.}, \bibinfo{author}{Sunehag, P.} \&
  \bibinfo{author}{Hutter, M.}
\newblock \bibinfo{title}{Feature reinforcement learning in practice}.
\newblock In \bibinfo{editor}{Sanner, S.} \& \bibinfo{editor}{Hutter, M.}
  (eds.) \emph{\bibinfo{booktitle}{Recent Advances in Reinforcement Learning}},
  vol. \bibinfo{volume}{7188} of \emph{\bibinfo{series}{Lecture Notes in
  Computer Science}}, \bibinfo{pages}{66--77} (\bibinfo{publisher}{Springer},
  \bibinfo{address}{Berlin, Germany}, \bibinfo{year}{2012}).

\bibitem{daswani2014feature}
\bibinfo{author}{Daswani, M.}, \bibinfo{author}{Sunehag, P.} \&
  \bibinfo{author}{Hutter, M.}
\newblock \bibinfo{title}{Feature reinforcement learning: State of the art}.
\newblock In \emph{\bibinfo{booktitle}{Proc. 28th {AAAI} Conf. Artif. Intell.:
  Sequential Decision Making with Big Data}}, \bibinfo{pages}{2--5}
  (\bibinfo{year}{2014}).

\bibitem{briegel2012projective}
\bibinfo{author}{Briegel, H.~J.} \& \bibinfo{author}{De~las Cuevas, G.}
\newblock \bibinfo{title}{Projective simulation for artificial intelligence}.
\newblock \emph{\bibinfo{journal}{Sci. Rep.}} \textbf{\bibinfo{volume}{2}},
  \bibinfo{pages}{400} (\bibinfo{year}{2012}).

\bibitem{Randomized_Algorithms_1995}
\bibinfo{author}{Motwani, R.} \& \bibinfo{author}{Raghavan, P.}
\newblock \emph{\bibinfo{title}{Randomized Algorithms}},
  chap.~\bibinfo{chapter}{6} (\bibinfo{publisher}{Cambridge University Press},
  \bibinfo{address}{New York, USA}, \bibinfo{year}{1995}).

\bibitem{Understanding_Intelligence_1999}
\bibinfo{author}{Pfeiffer, R.} \& \bibinfo{author}{Scheier, C.}
\newblock \emph{\bibinfo{title}{Understanding Intelligence}}
  (\bibinfo{publisher}{MIT Press}, \bibinfo{address}{Cambridge, MA, USA},
  \bibinfo{year}{1999}), \bibinfo{edition}{first} edn.

\bibitem{Childs:2003}
\bibinfo{author}{Childs, A.~M.} \emph{et~al.}
\newblock \bibinfo{title}{Exponential algorithmic speedup by a quantum walk}.
\newblock In \emph{\bibinfo{booktitle}{Proc. 35th Annu. ACM Symp. Theory
  Comput. (STOC)}}, \bibinfo{pages}{59--68} (\bibinfo{publisher}{ACM},
  \bibinfo{address}{New York, NY, USA}, \bibinfo{year}{2003}).

\bibitem{Kempe}
\bibinfo{author}{Kempe, J.}
\newblock \bibinfo{title}{Discrete quantum walks hit exponentially faster}.
\newblock \emph{\bibinfo{journal}{Probab. Theory Relat. Field}}
  \textbf{\bibinfo{volume}{133}}, \bibinfo{pages}{215--235}
  (\bibinfo{year}{2005}).

\bibitem{KMOR}
\bibinfo{author}{Krovi, H.}, \bibinfo{author}{Magniez, F.},
  \bibinfo{author}{Ozols, M.} \& \bibinfo{author}{Roland, J.}
\newblock \bibinfo{title}{Quantum walks can find a marked element on any
  graph}.
\newblock \emph{\bibinfo{journal}{Algorithmica}} \bibinfo{pages}{1--57}
  (\bibinfo{year}{2015}).

\bibitem{paparo2014quantum}
\bibinfo{author}{Paparo, G.~D.}, \bibinfo{author}{Dunjko, V.},
  \bibinfo{author}{Makmal, A.}, \bibinfo{author}{Martin-Delgado, M.~A.} \&
  \bibinfo{author}{Briegel, H.~J.}
\newblock \bibinfo{title}{Quantum speed-up for active learning agents}.
\newblock \emph{\bibinfo{journal}{Phys. Rev. X}} \textbf{\bibinfo{volume}{4}},
  \bibinfo{pages}{031002} (\bibinfo{year}{2014}).

\bibitem{dunjko2015quantum}
\bibinfo{author}{Dunjko, V.}, \bibinfo{author}{Friis, N.} \&
  \bibinfo{author}{Briegel, H.~J.}
\newblock \bibinfo{title}{Quantum-enhanced deliberation of learning agents
  using trapped ions}.
\newblock \emph{\bibinfo{journal}{New J. Phys.}} \textbf{\bibinfo{volume}{17}},
  \bibinfo{pages}{023006} (\bibinfo{year}{2015}).

\bibitem{friis2015coherent}
\bibinfo{author}{Friis, N.}, \bibinfo{author}{Melnikov, A.~A.},
  \bibinfo{author}{Kirchmair, G.} \& \bibinfo{author}{Briegel, H.~J.}
\newblock \bibinfo{title}{Coherent controlization using superconducting
  qubits}.
\newblock \emph{\bibinfo{journal}{Sci. Rep.}} \textbf{\bibinfo{volume}{5}},
  \bibinfo{pages}{18036} (\bibinfo{year}{2015}).

\bibitem{PhysRevLett.QML}
\bibinfo{author}{Dunjko, V.}, \bibinfo{author}{Taylor, J.~M.} \&
  \bibinfo{author}{Briegel, H.~J.}
\newblock \bibinfo{title}{Quantum-enhanced machine learning}.
\newblock \emph{\bibinfo{journal}{Phys. Rev. Lett.}}
  \textbf{\bibinfo{volume}{117}}, \bibinfo{pages}{130501}
  (\bibinfo{year}{2016}).
  
\bibitem{sriarunothai2017speedingup}
\bibinfo{author}{Sriarunothai, T.}, \bibinfo{author}{W{\"o}lk, S.}, \bibinfo{author}{Giri, G.~S.}, \bibinfo{author}{Friis, N.}, \bibinfo{author}{Dunjko, V.}, \bibinfo{author}{Briegel, H.~J.} \&
  \bibinfo{author}{Wunderlich, C.}
\newblock \bibinfo{title}{Speeding-up the decision making of a learning agent using an ion trap quantum processor}.
\newblock \emph{\bibinfo{journal}{arXiv:1709.01366}}  (\bibinfo{year}{2017}).

\bibitem{mautner2013projective}
\bibinfo{author}{Mautner, J.}, \bibinfo{author}{Makmal, A.},
  \bibinfo{author}{Manzano, D.}, \bibinfo{author}{Tiersch, M.} \&
  \bibinfo{author}{Briegel, H.~J.}
\newblock \bibinfo{title}{Projective simulation for classical learning agents:
  a comprehensive investigation}.
\newblock \emph{\bibinfo{journal}{New Gener. Comput.}}
  \textbf{\bibinfo{volume}{33}}, \bibinfo{pages}{69--114}
  (\bibinfo{year}{2015}).

\bibitem{melnikov2014projective}
\bibinfo{author}{Melnikov, A.~A.}, \bibinfo{author}{Makmal, A.} \&
  \bibinfo{author}{Briegel, H.~J.}
\newblock \bibinfo{title}{Projective simulation applied to the grid-world and
  the mountain-car problem}.
\newblock \emph{\bibinfo{journal}{arXiv:1405.5459}}  (\bibinfo{year}{2014}).

\bibitem{simon2016}
\bibinfo{author}{Hangl, S.}, \bibinfo{author}{Ugur, E.},
  \bibinfo{author}{Szedmak, S.} \& \bibinfo{author}{Piater, J.}
\newblock \bibinfo{title}{Robotic playing for hierarchical complex skill learning}.
\newblock In \emph{\bibinfo{booktitle}{Proc. IEEE/RSJ Int. Conf. Intell. Robots
  Syst.}}, \bibinfo{pages}{2799--2804} (\bibinfo{year}{2016}).
  
\bibitem{melnikov2017}
\bibinfo{author}{Melnikov, A. A.}, \bibinfo{author}{Poulsen Nautrup, H.},
  \bibinfo{author}{Krenn, M.},
  \bibinfo{author}{Dunjko, V.},
  \bibinfo{author}{Tiersch, M.},
  \bibinfo{author}{Zeilinger, A.} \& \bibinfo{author}{Briegel, H. J.}
\newblock \bibinfo{title}{Active learning machine learns to create new quantum experiments}.
\newblock \emph{\bibinfo{journal}{arXiv:1706.00868}}  (\bibinfo{year}{2017}).

\bibitem{bellman1957dynamic}
\bibinfo{author}{Bellman, R.~E.}
\newblock \emph{\bibinfo{title}{Dynamic Programming}}
  (\bibinfo{publisher}{Princeton University Press},
  \bibinfo{address}{Princeton, NJ, US}, \bibinfo{year}{1957}).

\bibitem{makmal2016meta}
\bibinfo{author}{Makmal, A.}, \bibinfo{author}{Melnikov, A.~A.},
  \bibinfo{author}{Dunjko, V.} \& \bibinfo{author}{Briegel, H.~J.}
\newblock \bibinfo{title}{Meta-learning within projective simulation}.
\newblock \emph{\bibinfo{journal}{IEEE Access}} \textbf{\bibinfo{volume}{4}},
  \bibinfo{pages}{2110--2122} (\bibinfo{year}{2016}).

\bibitem{Bandits05}
\bibinfo{author}{Wang, C.-C.}, \bibinfo{author}{Kulkarni, S.~R.} \&
  \bibinfo{author}{Poor, H.~V.}
\newblock \bibinfo{title}{Bandit problems with side observations}.
\newblock \emph{\bibinfo{journal}{{IEEE} Trans. Autom. Control}}
  \textbf{\bibinfo{volume}{50}}, \bibinfo{pages}{338--355}
  (\bibinfo{year}{2005}).

\bibitem{bjerland2015projective}
\bibinfo{author}{Bjerland, {\O}.~F.}
\newblock \emph{\bibinfo{title}{Projective simulation compared to reinforcement
  learning}}.
\newblock Master's thesis, \bibinfo{school}{Dept. Comput. Sci., Univ. Bergen},
  \bibinfo{address}{Bergen, Norway} (\bibinfo{year}{2015}).

\bibitem{tiersch2014adaptive}
\bibinfo{author}{Tiersch, M.}, \bibinfo{author}{Ganahl, E.~J.} \&
  \bibinfo{author}{Briegel, H.~J.}
\newblock \bibinfo{title}{Adaptive quantum computation in changing environments
  using projective simulation}.
\newblock \emph{\bibinfo{journal}{Sci. Rep.}} \textbf{\bibinfo{volume}{5}},
  \bibinfo{pages}{12874} (\bibinfo{year}{2015}).

\bibitem{simon2015evaluation}
\bibinfo{author}{Hangl, S.}
\newblock \emph{\bibinfo{title}{Evaluation and extensions of generalization in
  the projective simulation model}}.
\newblock \bibinfo{type}{Bachelor's thesis}, \bibinfo{school}{Univ. Innsbruck},
  \bibinfo{address}{Innsbruck, Austria} (\bibinfo{year}{2015}).

\end{thebibliography}
\end{document}